\documentclass[11pt,letterpaper]{article}
\usepackage[hyphens,spaces,obeyspaces]{url}
\usepackage[hyperref]{emnlp2017}
\usepackage{graphicx}
\usepackage{times}
\usepackage{latexsym}
\usepackage{amsmath}
\usepackage{amssymb}
\usepackage{array}

\usepackage{booktabs}
\usepackage{microtype}
\usepackage{cleveref}
\Crefformat{equation}{(#2#1#3)}

\newcommand{\ab}{\mathbf{a}}
\newcommand{\bb}{\mathbf{b}}
\newcommand{\vb}{\mathbf{v}}
\newcommand{\yhat}{\hat{y}}
\newcommand{\yb}{\mathbf{y}}

\newcommand{\RR}{\mathbb{R}}

\newcommand{\eat}[1]{\ignorespaces}

\newcolumntype{L}{>{\arraybackslash}m{7cm}}

\emnlpfinalcopy

\def\emnlppaperid{18}

\title{Neural Paraphrase Identification of Questions\\with Noisy Pretraining}

\author{Gaurav Singh Tomar \ \ Thyago Duque \ \ Oscar T{\"{a}}ckstr{\"{o}}m \\ {\bf Jakob Uszkoreit} \ \ {\bf Dipanjan Das}\\Google Inc.\\ \{\texttt{gtomar, duque, oscart, uszkoreit, dipanjand\}}\texttt{@google.com}}

\date{}

\begin{document}

\maketitle

\begin{abstract}
We present a solution to the problem of paraphrase identification of questions.
We focus on a recent dataset of question pairs annotated with binary paraphrase labels and show that a variant of the decomposable attention model \cite{parikh.etal.2016} results in accurate performance on this task, while being far simpler than many competing neural architectures.
Furthermore, when the model is pretrained on a noisy dataset of automatically collected question paraphrases, it obtains the best reported performance on the dataset.
\end{abstract}

\section{Introduction}
Question paraphrase identification is a widely useful NLP application.
For example, in question-and-answer (QA) forums ubiquitous on the Web, there are vast numbers of duplicate questions.
Identifying these duplicates and consolidating their answers increases the efficiency of such QA forums.
Moreover, identifying questions with the same semantic content could help Web-scale question answering systems that are increasingly concentrating on retrieving focused answers to users' queries.

Here, we focus on a recent dataset published by the QA website Quora.com containing over 400K annotated question pairs containing binary paraphrase labels.\footnote{See \url{https://data.quora.com/First-Quora-Dataset-Release-Question-Pairs}.}  We believe that this dataset presents a great opportunity to the NLP research community and practitioners due to its scale and quality; it can result in systems that accurately identify duplicate questions, thus increasing the quality of many QA forums.  We examine a simple model family, the \emph{decomposable attention model} of \newcite{parikh.etal.2016}, that has shown promise in modeling natural language inference and has inspired recent work on similar tasks \cite{ChenZLWJ16,kim:iclr2017}.

We present two contributions.  First, to mitigate data sparsity, we modify the input representation of the decomposable attention model to use sums of character $n$-gram embeddings instead of word embeddings.  We show that this model trained on the Quora dataset produces comparable or better results with respect to several complex neural architectures, all using pretrained word embeddings.  Second, to significantly improve our model performance, we pretrain \textit{all} our model parameters on the noisy, automatically collected question-paraphrase corpus Paralex \cite{paralex}, followed by fine-tuning the parameters on the Quora dataset.  This two-stage training procedure achieves the best result on the Quora dataset to date, and is also significantly better than learning \textit{only} the character $n$-gram embeddings during the pretraining stage.

\section{Related Work}\label{rel-work}
Paraphrase identification is a well-studied task in NLP \cite[\textit{inter alia}]{das2009paraphrase,chang2010discriminative,he-gimpel-lin:2015:EMNLP,wang-mi-ittycheriah:2016:COLING}.  Here, we focus on an instance, that of finding questions with identical meaning.  \newcite{lei-EtAl:2016:N16-1} consider a related task leveraging the AskUbuntu corpus \cite{dossantos-EtAl:2015:ACL-IJCNLP}, but it contains two orders of magnitude less annotations, thus limiting the quality of any model. Most relevant to this work is that of \newcite{wang:2017:ijcai}, who present the best results on the Quora dataset prior to this work. The \emph{bilateral multi-perspective matching} model (\textsc{BiMPM}) of \citeauthor{wang:2017:ijcai} uses a character-based LSTM \cite{hochreiter1997long} at its input representation layer, a layer of bi-LSTMs for computing context information, four different types of multi-perspective matching layers, an additional bi-LSTM aggregation layer, followed by a two-layer feedforward network for prediction.
In contrast, the decomposable attention model uses four simple feedforward networks to (self-)attend, compare and predict, leading to a more efficient architecture.
\textsc{BiMPM} falls short of our best performing model pretrained on noisy paraphrase data and uses more parameters than our best model.

Character-level modeling of text is a popular approach.
While conceptually simple, character $n$-gram embeddings are a highly competitive representation \cite{huang-deep-structured-semantic,wieting-EtAl:2016:EMNLP2016,DBLP:journals/corr/BojanowskiGJM16}.
More complex representations built directly from individual characters have also been proposed \cite{sennrich-haddow-birch:2016:P16-12,luong-manning:2016:P16-1,Kim:2016:CNL:3016100.3016285,chung-cho-bengio:2016:P16-1,ling-EtAl:2015:EMNLP2}.
These representations are robust to out-of-vocabulary items, often producing improved results.  Our pretraining procedure is reminiscent of several recent papers \cite[\textit{inter alia}]{wieting-EtAl:2016:EMNLP2016} who aim for general purpose character $n$-gram embeddings.  In contrast, we pretrain all model parameters on automatic but in-domain paraphrase data.  We employ the same neural architecture as our end task, similar to prior work on multi-task learning \cite[\textit{inter alia}]{sogaard-goldberg:2016:P16-2}, but use a simpler learning setup.

\section{Approach}
Our starting point is the decomposable attention model \cite[\textsc{DecAtt} henceforth]{parikh.etal.2016}, which despite its simplicity and efficiency has been shown to work remarkably well for the related task of natural language inference \cite{bowman2015large}.
We extend this model with character $n$-gram embeddings and noisy pretraining for the task of question paraphrase identification.

\subsection{Problem Formulation}\label{subsec:prob_formulation}
Let $\ab = (a_1,\ldots,a_{\ell_a})$ and $\bb = (b_1,\ldots,b_{\ell_b})$ be two input texts consisting of $\ell_a$ and $\ell_b$ tokens, respectively.
We assume that each $a_i$, $b_j \in \RR^d$ is encoded in a vector of dimension $d$.
A context window of size $c$ is subsequently applied, such that the input to the model $(\bar \ab, \bar \bb)$ consists of partly overlapping phrases $\bar{a}_i = [a_{i-c}, \ldots, a_{i}, \ldots , a_{i+c}], \bar{b}_j = [b_{j-c}, \ldots, b_{j}, \ldots , b_{j+c}] \in \RR^{2c+1 \times d}$.
The model is estimated using training data in the form of labeled pairs $\{\ab^{(n)}, \bb^{(n)}, \yb^{(n)} \}_{n=1}^N$, where $\yb^{(n)} \in \{0, 1\}$ is a binary label indicating whether $\ab$ is a paraphrase of $\bb$ or not.
Our goal is to predict the correct label $\yb$ given a pair of previously unseen texts $(\ab, \bb)$.

\subsection{The Decomposable Attention Model}
The \textsc{DecAtt} model divides the prediction into three steps: \emph{Attend}, \emph{Compare} and \emph{Aggregate}. Due to lack of space, we only provide a brief outline below and refer to \newcite{parikh.etal.2016} for further details on each of these steps. 

\paragraph{Attend.}
First, the elements of $\bar{\ab}$ and $\bar{\bb}$ are aligned using a variant of neural attention \cite{bahdanau2014neural} to decompose the problem into the comparison of aligned phrases.
\begin{align}\label{eq:unnormalized_attention}
e_{ij} := F(\bar{a}_i)^\top F(\bar{b}_j)\,.
\end{align}
The function $F$ is a feedforward network.  The aligned phrases are computed as follows:
\begin{align}\label{eq:normalized_attention}
\beta_i &:= \sum_{j=1}^{\ell_b} \frac{\exp(e_{ij})}{\sum_{k=1}^{\ell_b} \exp(e_{ik})} \bar{b}_j\,,  \notag \\
\alpha_j &:= \sum_{i=1}^{\ell_a} \frac{\exp(e_{ij})}{\sum_{k=1}^{\ell_a} \exp(e_{kj})} \bar{a}_i\,.
\end{align}
Here $\beta_i$ is the subphrase in $\bar{\bb}$ that is (softly) aligned to $\bar{a}_i$ and vice versa for $\alpha_j$.
Optionally, the inputs $\bar{\ab}$ and $\bar{\bb}$ to \Cref{eq:unnormalized_attention} can be replaced by input representations passed through a ``self-attention" step to capture longer context.
In this optional step, we modify the input representations using ``self-attention" to encode compositional relationships between words within each sentence, as proposed by \cite{cheng-dong-lapata:2016:EMNLP2016}.
Similar to \Cref{eq:unnormalized_attention}, we define
\begin{align}\label{eq:unnormalized_self_attention}
f_{ij} := F_{self}(\bar{a}_i)^\top F'_{self}(\bar{a}_j)\,.
\end{align}
The function $F_{self}$ and $F'_{self}$ are feedforward networks. The self-aligned phrases are then computed as follows:
\begin{align}\label{eq:normalized_slef_attention}
a_i' &:= \sum_{j=1}^{\ell_a} \frac{\exp(f_{ij} + d_{i-j})}{\sum_{k=1}^{\ell_a} \exp(f_{ik} + d_{i-k})} a_j\,.
\end{align}
where $d_{i-j}$ is a learned distance-sensitive bias term. Subsequent steps then use modified input representations defined as $\bar{a}_i$ := [$a_i$,$a_i'$] and $\bar{b}_i$ := [$b_i$,$b_i'$].
\paragraph{Compare.}  Second, we separately compare the aligned phrases $\{(\bar{a}_i, \beta_i)\}_{i=1}^{\ell_a}$ and $\{(\bar{b}_j, \alpha_j)\}_{j=1}^{\ell_b}$ using a feedforward network $G$:
\begin{align}
\vb_{1,i} &:= G([\bar{a}_i, \beta_i])\quad \forall i \in \langle 1,\ldots, \ell_a\rangle\,, \notag \\
\vb_{2,j} &:= G([\bar{b}_j, \alpha_j])\quad \forall j \in \langle 1,\ldots, \ell_b\rangle\,.
\label{eq:gdefine}
\end{align}
where the brackets $[\cdot, \cdot]$ denote concatenation. 

\paragraph{Aggregate.} Finally, the sets $\{\vb_{1,i}\}_{i=1}^{\ell_a}$ and $\{\vb_{2,j}\}_{j=1}^{\ell_b}$ are aggregated by summation. The sum of two sets is concatenated and passed through another feedforward network followed by a linear layer, to predict the label $\yhat$.

\subsection{Character $n$-Gram Word Encodings}
Parikh~et~al.\nocite{parikh.etal.2016} assume that each token $a_i$, $b_j \in \RR^d$ is directly embedded in a vector of dimension $d$; in practice, they used pretrained word embeddings.
Inspired by prior work mentioned in \Cref{rel-work}, we use an alternative approach and instead represent each token as a sum of its embedded character $n$-grams. This allows for more effective parameter sharing at a small additional computational cost.  As observed in \Cref{sec:experiments}, this leads to better results compared to word embeddings.

\subsection{Noisy Pretraining}
While character $n$-gram encodings help in effective parameter sharing, data sparsity remains an issue.
Pretraining embeddings with a task-agnostic objective on large-scale corpora \cite{pennington2014glove} is a common remedy to this problem.
However, such pretraining is limited in the following ways. First, it only applies to the input representation, leaving subsequent parts of the model to random initialization. Second, there may be a domain mismatch unless embeddings are pretrained on the same domain as the end task (e.g., questions). Finally, since the objective used for pretraining differs from that of the end task (e.g., paraphrase identification), the embeddings may be suboptimal.

As an alternative to task-agnostic pretraining of embeddings on a very large corpus, we propose to pretrain all parameters of the model on a modest-sized corpus of automatically gathered, and therefore noisy examples, drawn from a similar domain.\footnote{Paralex is gathered from WikiAnswers, a QA forum.}
As observed in \Cref{sec:experiments}, such noisy pretraining of the full model results in more accurate performance compared to using pretrained embeddings, as well as compared to only pretraining embeddings on the noisy in-domain corpus.\footnote{The Quora data is similar to the Paralex corpus and we exploit this by pretraining our entire model on the latter.  It can be argued that not all sentence pair modeling tasks may benefit similarly from the Paralex corpus and a detailed empirical study is warranted to investigate that; in this work, we restrict our scope to only the question paraphrase identification task, a very useful NLP application by itself.}

\begin{figure}[!t]
\centering
\includegraphics[width=0.8\columnwidth]{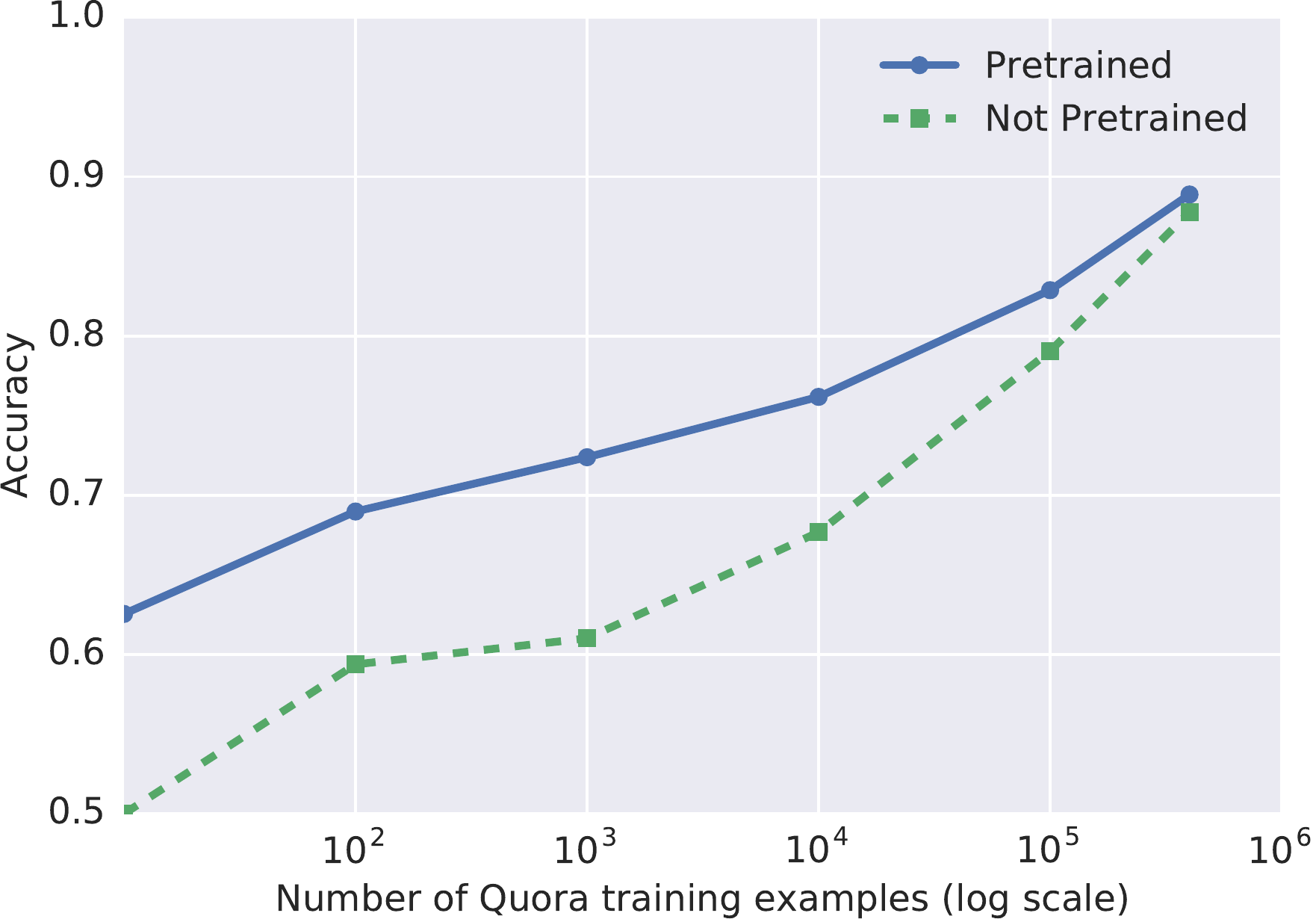}
\caption{\label{fig:learning-curve}
Learning curves for the Quora development set with and without pretraining on Paralex.}
\vspace{-0.2cm}
\end{figure}

\begin{table*}[!th]
\begin{center}
\resizebox{\textwidth}{!}{
\begin{tabular}{cLLcccc}
\toprule
ID & Question 1 & Question 2 & \textsc{DecAtt}$_\mathrm{glove}$ & \textsc{DecAtt}$_\mathrm{char}$ & pt-\textsc{DecAtt}$_\mathrm{char}$ & Gold\\
\midrule
A & How shall I start my preparation for IIT-JEE in class 10? & Should I start preparing for the IIT JEE in class 10 only? & N & \textbf{Y} & \textbf{Y} & \textbf{Y}\\[0.3cm]
B & What is fama french three factor model? & What is Fama-French three factor model? & N & \textbf{Y} & \textbf{Y} & \textbf{Y}\\[0.3cm]
\midrule
C & How does PayPal work in India? & Does PayPal work in India? What features of PayPal are available in India? & Y & Y & \textbf{N} & \textbf{N}\\[0.3cm]
D & What are the similarities between British English and American English? & What are the similarities between American English and British English? & N & N & \textbf{Y} & \textbf{Y}\\[0.3cm]
\midrule
E & How is buying land on the moon a good investment? Why do people buy land on the moon? & At \$20 an acre, isn't buying moon plots a solid investment? & N &  N & N & \textbf{Y}\\[0.3cm]
F & What can wrestlers do to prevent cauliflower ears? & Why do wrestlers have deformed ears? & N & N & N & \textbf{Y}\\
\bottomrule
\end{tabular}}
\end{center}
\caption{Example wins and losses from the \textsc{DecAtt}$_\mathrm{glove}$, \textsc{DecAtt}$_\mathrm{char}$ and the pt-\textsc{DecAtt}$_\mathrm{char}$ models.}
\label{table:error-analysis}

\end{table*}

\begin{table}[!t]

\centering
\begin{tabular*}{\columnwidth}{lccr}
\toprule
Method & Dev Acc & Test Acc\\
\midrule
Siamese-CNN & - & 79.60\\
Multi-Perspective CNN & - & 81.38\\
Siamese-LSTM & - & 82.58\\
Multi-Perspective-LSTM & - & 83.21\\
L.D.C & - & 85.55\\
\textsc{BiMPM} & 88.69 & 88.17\\
\midrule
\textsc{FFNN}$_\mathrm{word}$ &  85.07 & 84.35 &  \\
\textsc{FFNN}$_\mathrm{char}$ & 86.01 & 85.06\\
\midrule
\textsc{DecAtt}$_\mathrm{word}$ & 86.04 & 85.27 \\
\textsc{DecAtt}$_\mathrm{glove}$ & 87.42 & 86.52 \\
\textsc{DecAtt}$_\mathrm{char}$ & 87.78 & 86.84\\
\textsc{DecAtt}$_\mathrm{paralex-char}$ & 87.80 & 87.77\\
\midrule
pt-\textsc{DecAtt}$_\mathrm{word}$ & 88.44 & 87.54\\
pt-\textsc{DecAtt}$_\mathrm{char}$ & \bf{88.89} & \bf{88.40}\\
\bottomrule
\end{tabular*}
\caption{Results on the Quora development and test sets in terms of accuracy.  The first six rows are taken from \cite{wang:2017:ijcai}.}
\label{table:quora-results}
\end{table}

\section{Experiments}\label{sec:experiments}

\subsection{Implementation Details}
\noindent \textbf{Datasets}  We evaluate our models on the Quora question paraphrase dataset which contains over 400,000 question pairs with binary labels.  We use the same data and split as \newcite{wang:2017:ijcai}, with 10,000 question pairs each for development and test, who also provide preprocessed and tokenized question pairs.\footnote{This split is available at \url{https://zhiguowang.github.io}.}  We duplicated the training set, which has approximately 36\% positive and 64\% negative pairs, by adding question pairs in reverse order (since our model is not symmetric). When pretraining the full model parameters, we use the Paralex corpus \cite{paralex}, which consists of 36 million noisy paraphrase pairs including duplicate reversed paraphrases.  We created 64 million artificial negative paraphrase pairs (reflecting the class balance of the Quora training set) by combining the following three types of negatives in equal proportions: (1) random unrelated questions, (2) random questions that share a single word, and (3) random questions that share all but one word.\footnote{More complex sampling procedures are possible, for example, by using pretrained word embeddings.}

\noindent \textbf{Hyperparameters} 
We tuned the following hyperparameters by grid search on the development set (settings for our best model are in parenthesis): embedding dimension (300), shape of all feedforward networks (two layers with 400 and 200 width), character $n$-gram sizes (5), context size (1), learning rate (0.1 for both pretraining and tuning), batch size (256 for pretraining and 64 for tuning), dropout ratio (0.1 for tuning) and prediction threshold (positive paraphrase for a score $\ge$ 0.3).  We examined whether self-attention helps or not for all model variants, and found that it does for our best model.
Note that we tried multiple orders of character n-grams with n $\in$ \{3, 4, 5\} both individually and separately but 5-grams alone worked better than these alternatives.

\noindent \textbf{Baselines} We implemented several baseline models.  In our first two baselines, each question is represented by concatenating the sum of its unigram word embeddings and the sum of its trigram vectors, where each trigram vector is a concatenation of 3 adjacent word embeddings.  The two question representations are then concatenated and fed to a feedforward network of shape [800, 400, 200].  We call these \textsc{FFNN}$_\mathrm{word}$ and \textsc{FFNN}$_\mathrm{char}$; in the latter, word embeddings are just sums of character $n$-gram embeddings.  Second, we compare purely supervised variants of decomposable attention model, namely a word-based model without any pretrained embeddings (\textsc{DecAtt}$_\mathrm{word}$), a word-based model with GloVe \cite{pennington2014glove} embeddings (\textsc{DecAtt}$_\mathrm{glove}$), a character $n$-gram model (\textsc{DecAtt}$_\mathrm{char}$)  without pretrained embeddings and \textsc{DecAtt}$_\mathrm{paralex-char}$ whose character $n$-gram embeddings are pretrained with Paralex while all other parameters are learned from scratch on Quora.  Finally we present a baseline where a word-based model is pretrained completely on Paralex (pt-\textsc{DecAtt}$_\mathrm{word}$) and our best model which is a character $n$-gram model pretrained completely on Paralex  (pt-\textsc{DecAtt}$_\mathrm{char}$). 
Note that in case of character $n$-gram based models, for tokens shorter than n characters, we backoff and emit the token itself. Also, boundary markers were added at the beginning and end of each word. 

\subsection{Results}
Other than our baselines, we compare with \newcite{wang:2017:ijcai} in \Cref{table:quora-results}. We observe that the simple FFNN baselines work better than more complex Siamese and Multi-Perspective CNN or LSTM models, more so if character $n$-gram based embeddings are used. Our basic decomposable attention model \textsc{DecAtt}$_\mathrm{word}$ without pre-trained embeddings is better than most of the models, all of which used GloVe embeddings. An interesting observation is that  \textsc{DecAtt}$_\mathrm{char}$ model without any pretrained embeddings outperforms \textsc{DecAtt}$_\mathrm{glove}$ that uses task-agnostic GloVe embeddings. Furthermore, when character $n$-gram embeddings are pre-trained in a task-specific manner in \textsc{DecAtt}$_\mathrm{paralex-char}$ model, we observe a significant boost in performance. \footnote{Note that Paralex is orders of magnitude smaller than the corpus used to pretrain GloVe.}

The final two rows of the table show results achieved by pt-\textsc{DecAtt}$_\mathrm{word}$ and pt-\textsc{DecAtt}$_\mathrm{char}$.  We note that the former falls short of the \textsc{DecAtt}$_\mathrm{paralex-char}$, which shows that character $n$-gram representations are powerful.  Finally, we note that our best performing model is pt-\textsc{DecAtt}$_\mathrm{char}$, which leverages the full power of character embeddings and pretraining the model on Paralex.

Noisy pretraining gives more significant gains in case of smaller human annotated data as can be seen in \Cref{fig:learning-curve} where non-pretrained  \textsc{DecAtt}$_\mathrm{char}$ and pretrained pt-\textsc{DecAtt}$_\mathrm{char}$ are compared on a logarithmic scale of number of Quora examples. It also gives an important insight into trade off between having more but costly human annotated data versus cheap but noisy pretraining.  \Cref{table:error-analysis} shows some example predictions from the  \textsc{DecAtt}$_\mathrm{glove}$, \textsc{DecAtt}$_\mathrm{char}$ and the pt-\textsc{DecAtt}$_\mathrm{char}$ models.  The GloVe-trained model often makes mistakes related to spelling and tokenization artifacts.  We observed that hyperparameter tuning resulted in settings where non-pretrained models did not use self-attention while the pretrained character based model did, thus learning better long term context at its input layer; this is reflected in example D which shows an alternation that our best model captures.  Finally, E and F show pairs that present complex paraphrases that none of our models capture.

\section{Conclusion and Future Work}
We presented a focused contribution on question paraphrase identification, on the recently published Quora corpus.
First, we showed that replacing the word embeddings of the decomposable attention model of \newcite{parikh.etal.2016} with character $n$-gram embeddings results in significantly better accuracy on this task. Second, we showed that pretraining the full model on automatically labeled noisy, but task-specific data results in further improvements.
Our methods perform better than several complex neural architectures and achieve state of the art. While conceptually simple, we believe that these are two important insights that may be more widely applicable within the field of natural language understanding. We leave investigation of this claim to future work that may involve evaluation on related tasks such as recognizing textual entailment.
\bibliography{emnlp2017}

\begin{thebibliography}{}
\expandafter\ifx\csname natexlab\endcsname\relax\def\natexlab#1{#1}\fi

\bibitem[{Bahdanau et~al.(2015)Bahdanau, Cho, and Bengio}]{bahdanau2014neural}
Dzmitry Bahdanau, Kyunghyun Cho, and Yoshua Bengio. 2015.
\newblock Neural machine translation by jointly learning to align and
  translate.
\newblock In {\em Proceedings of ICLR\/}.

\bibitem[{Bojanowski et~al.(2016)Bojanowski, Grave, Joulin, and
  Mikolov}]{DBLP:journals/corr/BojanowskiGJM16}
Piotr Bojanowski, Edouard Grave, Armand Joulin, and Tomas Mikolov. 2016.
\newblock Enriching word vectors with subword information.
\newblock {\em {arXiv}\/} 1607.04606.

\bibitem[{Bowman et~al.(2015)Bowman, Angeli, Potts, and
  Manning}]{bowman2015large}
Samuel~R. Bowman, Gabor Angeli, Christopher Potts, and Christopher~D. Manning.
  2015.
\newblock A large annotated corpus for learning natural language inference.
\newblock In {\em Proceedings of EMNLP\/}.

\bibitem[{Chang et~al.(2010)Chang, Goldwasser, Roth, and
  Srikumar}]{chang2010discriminative}
Ming-Wei Chang, Dan Goldwasser, Dan Roth, and Vivek Srikumar. 2010.
\newblock Discriminative learning over constrained latent representations.
\newblock In {\em Proceedings of HLT-NAACL\/}.

\bibitem[{Chen et~al.(2016)Chen, Zhu, Ling, Wei, and Jiang}]{ChenZLWJ16}
Qian Chen, Xiaodan Zhu, Zhen{-}Hua Ling, Si~Wei, and Hui Jiang. 2016.
\newblock Enhancing and combining sequential and tree {LSTM} for natural
  language inference.
\newblock {\em arXiv 1609.06038\/} .

\bibitem[{Cheng et~al.(2016)Cheng, Dong, and
  Lapata}]{cheng-dong-lapata:2016:EMNLP2016}
Jianpeng Cheng, Li~Dong, and Mirella Lapata. 2016.
\newblock \href{https://aclweb.org/anthology/D16-1053}{Long short-term
  memory-networks for machine reading}.
\newblock In {\em Proceedings of the 2016 Conference on Empirical Methods in
  Natural Language Processing\/}. Association for Computational Linguistics,
  Austin, Texas, pages 551--561.
\newblock
  \href{https://aclweb.org/anthology/D16-1053}{https://aclweb.org/anthology/D16-1053}.

\bibitem[{Chung et~al.(2016)Chung, Cho, and
  Bengio}]{chung-cho-bengio:2016:P16-1}
Junyoung Chung, Kyunghyun Cho, and Yoshua Bengio. 2016.
\newblock A character-level decoder without explicit segmentation for neural
  machine translation.
\newblock In {\em Proceedings of ACL\/}.

\bibitem[{Das and Smith(2009)}]{das2009paraphrase}
Dipanjan Das and Noah~A. Smith. 2009.
\newblock Paraphrase identification as probabilistic quasi-synchronous
  recognition.
\newblock In {\em Proceedings of ACL-IJCNLP\/}.

\bibitem[{dos Santos et~al.(2015)dos Santos, Barbosa, Bogdanova, and
  Zadrozny}]{dossantos-EtAl:2015:ACL-IJCNLP}
Cicero dos Santos, Luciano Barbosa, Dasha Bogdanova, and Bianca Zadrozny. 2015.
\newblock Learning hybrid representations to retrieve semantically equivalent
  questions.
\newblock In {\em Proceedings of ACL\/}.

\bibitem[{Fader et~al.(2013)Fader, Zettlemoyer, and Etzioni}]{paralex}
Anthony Fader, Luke Zettlemoyer, and Oren Etzioni. 2013.
\newblock Paraphrase-driven learning for open question answering.
\newblock In {\em Proceedings of ACL\/}.

\bibitem[{He et~al.(2015)He, Gimpel, and Lin}]{he-gimpel-lin:2015:EMNLP}
Hua He, Kevin Gimpel, and Jimmy Lin. 2015.
\newblock Multi-perspective sentence similarity modeling with convolutional
  neural networks.
\newblock In {\em Proceedings of EMNLP\/}.

\bibitem[{Hochreiter and Schmidhuber(1997)}]{hochreiter1997long}
Sepp Hochreiter and J{\"u}rgen Schmidhuber. 1997.
\newblock Long short-term memory.
\newblock {\em Neural Computation\/} 9(8):1735--1780.

\bibitem[{Huang et~al.(2013)Huang, He, Gao, Deng, Acero, and
  Heck}]{huang-deep-structured-semantic}
Po-Sen Huang, Xiaodong He, Jianfeng Gao, Li~Deng, Alex Acero, and Larry Heck.
  2013.
\newblock Learning deep structured semantic models for web search using
  clickthrough data.
\newblock In {\em Proceedings of {CIKM}\/}.

\bibitem[{Kim et~al.(2017)Kim, Denton, Hoang, and Rush}]{kim:iclr2017}
Yoon Kim, Carl Denton, Loung Hoang, and Alexander~M. Rush. 2017.
\newblock Neural machine translation by jointly learning to align and
  translate.
\newblock In {\em Proceedings of ICLR\/}.

\bibitem[{Kim et~al.(2016)Kim, Jernite, Sontag, and
  Rush}]{Kim:2016:CNL:3016100.3016285}
Yoon Kim, Yacine Jernite, David Sontag, and Alexander~M. Rush. 2016.
\newblock Character-aware neural language models.
\newblock In {\em Proceedings of AAAI\/}.

\bibitem[{Lei et~al.(2016)Lei, Joshi, Barzilay, Jaakkola, Tymoshenko,
  Moschitti, and M\`{a}rquez}]{lei-EtAl:2016:N16-1}
Tao Lei, Hrishikesh Joshi, Regina Barzilay, Tommi Jaakkola, Kateryna
  Tymoshenko, Alessandro Moschitti, and Llu\'{i}s M\`{a}rquez. 2016.
\newblock Semi-supervised question retrieval with gated convolutions.
\newblock In {\em Proceedings of NAACL\/}.

\bibitem[{Ling et~al.(2015)Ling, Dyer, Black, Trancoso, Fermandez, Amir,
  Marujo, and Luis}]{ling-EtAl:2015:EMNLP2}
Wang Ling, Chris Dyer, Alan~W Black, Isabel Trancoso, Ramon Fermandez, Silvio
  Amir, Luis Marujo, and Tiago Luis. 2015.
\newblock Finding function in form: Compositional character models for open
  vocabulary word representation.
\newblock In {\em Proceedings of EMNLP\/}.

\bibitem[{Luong and Manning(2016)}]{luong-manning:2016:P16-1}
Minh-Thang Luong and Christopher~D. Manning. 2016.
\newblock Achieving open vocabulary neural machine translation with hybrid
  word-character models.
\newblock In {\em Proceedings of ACL\/}.

\bibitem[{Parikh et~al.(2016)Parikh, T\"{a}ckstr\"{o}m, Das, and
  Uszkoreit}]{parikh.etal.2016}
Ankur Parikh, Oscar T\"{a}ckstr\"{o}m, Dipanjan Das, and Jakob Uszkoreit. 2016.
\newblock A decomposable attention model for natural language inference.
\newblock In {\em Proceedings of EMNLP\/}.

\bibitem[{Pennington et~al.(2014)Pennington, Socher, and
  Manning}]{pennington2014glove}
Jeffrey Pennington, Richard Socher, and Christopher~D. Manning. 2014.
\newblock {GloVe}: Global vectors for word representation.
\newblock In {\em Proceedings of EMNLP\/}.

\bibitem[{Sennrich et~al.(2016)Sennrich, Haddow, and
  Birch}]{sennrich-haddow-birch:2016:P16-12}
Rico Sennrich, Barry Haddow, and Alexandra Birch. 2016.
\newblock Neural machine translation of rare words with subword units.
\newblock In {\em Proceedings of ACL\/}.

\bibitem[{S{\o}gaard and Goldberg(2016)}]{sogaard-goldberg:2016:P16-2}
Anders S{\o}gaard and Yoav Goldberg. 2016.
\newblock Deep multi-task learning with low level tasks supervised at lower
  layers.
\newblock In {\em Proceedings of ACL\/}.

\bibitem[{Wang et~al.(2017)Wang, Hamza, and Florian}]{wang:2017:ijcai}
Zhiguo Wang, Wael Hamza, and Radu Florian. 2017.
\newblock Bilateral multi-perspective matching for natural language sentences.
\newblock In {\em Proceedings of IJCAI\/}.

\bibitem[{Wang et~al.(2016)Wang, Mi, and
  Ittycheriah}]{wang-mi-ittycheriah:2016:COLING}
Zhiguo Wang, Haitao Mi, and Abraham Ittycheriah. 2016.
\newblock Sentence similarity learning by lexical decomposition and
  composition.
\newblock In {\em Proceedings of {COLING}\/}.

\bibitem[{Wieting et~al.(2016)Wieting, Bansal, Gimpel, and
  Livescu}]{wieting-EtAl:2016:EMNLP2016}
John Wieting, Mohit Bansal, Kevin Gimpel, and Karen Livescu. 2016.
\newblock Charagram: Embedding words and sentences via character n-grams.
\newblock In {\em Proceedings of EMNLP\/}.

\end{thebibliography}
\bibliographystyle{emnlp_natbib}

\end{document}